\documentclass[runningheads]{llncs}

\usepackage{graphicx}
\usepackage{amsmath,amssymb}
\usepackage{algorithmic,algorithm}
\usepackage{multicol,multirow}
\usepackage{color}
\usepackage{navigator}

\newif\ifBibtex\Bibtexfalse

\newif\ifDraft\Draftfalse

\newcommand{\secref}[1]{Section \ref{#1}}
\newcommand{\figref}[1]{Fig.~\ref{#1}}
\newcommand{\tabref}[1]{Table \ref{#1}}

\ifDraft

\else

\fi

\newcommand{\draftcontent}[1]{}

\begin{document}
\title{Deep Representation Learning with an Information-theoretic Loss}

%
%
\author{Shin Ando\inst{1} 
 }
\authorrunning{\phantom{SA}}
%
\institute{Department of Business Economics\\
School of Management\\
Tokyo University of Science\\
 \email{ando@rs.tus.ac.jp}%
}
\maketitle              
\begin{abstract}
This paper proposes a deep representation learning using an information-theoretic loss with an explicit focus on increasing the inter-class distances as well as within-class similarity in the embedded space.  
Tasks such as anomaly and out-of-distribution detection, in which test samples comes from classes unseen in training, are problematic for deep neural networks. For such tasks, it is not sufficient to merely discriminate between known classes. Our intuion is that by mapping the known classes to compact and disperse regions, the possibility of unseen and known overlapping in the embedded space may be reduced.
We derive a loss from Information Bottleneck principle, which reflects the inter-class distances as well as the compactness within classes, thus will extend the existing deep data description models. 
Our empirical study shows that the proposed model improves the dispersion of normal classes in the embedded space, and subsequently contributes to improved detection out-of-distribution samples.

\end{abstract}

\section{Introduction}
Recently, problems of out-of-distribution detection \cite{yang2021generalized} and anomaly detection \cite{1541882} have attracted interests in studies of deep neural networks (DNNs). It is a practical issue that DNNs show unexpected behaviors when testing samples come from unseen classes in training. DNN classifiers typically can show high confidence when predicting such samples as those from a known class. 
This behavior can be attributed to the discriminative softmax classifier in DNNs \cite{10.1145/3394486.3403189}. It has been proposed that deep data description models \cite{pmlr-v80-ruff18a}, which describes the normal data with a hypersphere, and multi-class data description (MCDD) which describes known classes  as Gaussian components, respectively in the embedded space, can achieve better detection performances for outlier and OOD detection tasks.
Both data description models define their losses such that each data is projected onto the proximity of a class center in the embedded space. Subsequently, they can identify a test sample outside of the hypersphere or has a low probability over all known classes as an outlier or an OOD sample.  

In above settings, the anomalies and the out-of-distribution samples are not available in training, making an attempt to learn the separation between them and known class distributions directly infeasible. A practical approach, instead, is to a) enclose each known class in a compact region and b) separate them from each other as much as possible so as to expand the space in between that yield low probabilities over known classes.

Previous studies, namely data description models, employed both max-margin and MAP losses, which placed emphases on a), the compactness of each class individually.
However, it is our intuition that consideration of inter-class separation can have substantial impact, as the it introduces supervising information regarding numerous combinations of heterogeneous class pairs. 

In this paper, we present an information-theoretic loss based on the information bottleneck principle\cite{DBLP:journals/corr/physics-0004057}, from 
which we can derive the relation between a) the intra-class similarity and b) the inter-class discrepancies. 


In our empirical study, we setup an out-of-distribution detection task using the MNIST dataset. The proposed model yields high detection performance and its graphical analysis shows that the proposed model contributes to the disperse mapping of normal classes.

This paper is organized as follows.  \secref{sec:Related_Work} describes the related studies, and \secref{sec:XDD} describes the analysis of the information bottleneck loss.
\secref{sec:experiments} presents the setup and the results of the empirical study. We state our conclusion in \secref{sec:conclusion}. 

\section{Related Work} \label{sec:Related_Work}

\subsection{Deep Data Description}\label{subsec:DFDD}
The support vector data description \cite{Tax:2004:SVD:960091.960109} aims to find a spherical decision boundary that encloses the normal data, in order to detect whether a new sample comes from the same distribution or is an outlier.
Deep-SVDD \cite{pmlr-v80-ruff18a} has employed the embedding functions of deep neural networks (DNN) to capture the structure of normal data.

Deep Multi-class Data Description (MCDD) \cite{10.1145/3394486.3403189} was introduced as an extension of the Deep SVDD for out-of-distribution (OOD) detection.
A DNN is trained such that the embedding function $f$ maps the labeled data onto the close proximity of the centers of the corresponding class $k$, to find Gaussian components which describes the training data in the embedded space ${\mathcal Z}$.

Describing the $k^\text{th}$ component as a multinormal distribution   
\begin{equation}
P(z|y=k) = {\mathcal N}\left(f(z;W)|\mu_k,\sigma_k^2I\right)
\label{eq:P(z|y=k)}
\end{equation}

The Deep MCDD loss ${\mathcal L}_\text{MCDD}$ is defined as a MAP loss of the generative classifier as 
\begin{eqnarray}
{\mathcal L}_\text{MCDD}&=&-\frac{1}{N}\sum_{i=1}^N\log
\frac{P(y=k)P(x|y=k)}{\sum_{k'}P(y=k')P(x|y=k')}
\nonumber\\
&=&\frac{1}{N}\sum_{i=1}^N\log\frac{\exp\left(-D_{y_i}(x_i)+b_{y_i}\right)}%
{\sum_{k=1}^K\exp\left(-D_k(x_i)+b_k\right)}
\label{eq:L_MCDD}
\end{eqnarray}
where $D_k(x)$ is the distance from the class centers given \eqref{eq:P(z|y=k)}.
\begin{equation}
D_k(x)
\approx \frac{\|f(x;W)-\mu_k\|^2}{2\sigma^2_k} + \log\sigma_k^d
\label{eq:D_k(x)=} 
\end{equation} 

From equations \ref{eq:L_MCDD} and \ref{eq:D_k(x)=},  
the Deep MCDD training can be considered a minimization of the intra-class deviation in the embedded space.

\section{Analysis of the Information Bottleneck Loss}\label{sec:XDD}
\subsection{Derivation}
The information bottleneck \cite{DBLP:journals/corr/physics-0004057,DBLP:conf/iclr/AlemiFD017}
 is a principle for extracting relevant information in the input variable $X$ about the output variable $Y$. The mutual information $I(\cdot;\cdot)$ quantifies the statistical dependence between the two variables. 
We attempt to learn a compressed representation of $X$, denoted as $Z$, by discarding irrelevant information that do not contribute to the prediction of $Y$.
 
In \cite{7133169}, it was proposed that Information Bottleneck principle may be used for layer-wise analyses of DNNs in terms of the efficiency of compression. We, however, focus on utilizing the above rate-distortion function for training a DNN. 
In this paper, therefore, we consider ${Z}$ to be a function of $X$, such that $f:{\mathcal X}\to{\mathcal Z}\subset{\mathbb R}^d$, and $f$ the embedding layers of a trained DNN model.
${\mathcal X, Y, Z}$ denotes the subspaces of data, label, and the embedded features, from which variables $X, Y, Z$ take their values, respectively. 

Finding an optimal $Z$ leads to a minimization problem for a Lagrangian  
\begin{equation}
{\mathcal L} = I(X;Z) - \beta I(Z;Y) 
\label{eq:L=I(X;Z)}
\end{equation}
This problem is referred to as a rate-distortion problem, as $I(X;Z)$ is the amount information maintained in $Z$ and a measure of the compression rate, while $I(Y;Z)$ is the amount of relevant information about $Y$ thus a measure of the distortion.
The Lagrangian multiplier $\beta$ represents the trade-off between the two terms.

The mutual information can be rewritten as the Kullback-Leibler divergence between the marginal and the conditional probability distributions 
\begin{equation}
I(X;Z) = \mathbb{E}_{x,z}\left[ D_\text{KL}\left(p(z|x)\|p(z)\right)\right]
\label{eq:I(X;Z)=D_KL}
\end{equation}

We model the empirical distribution of $p(z)$ by the average of the Dirac delta functions, 
\begin{equation}
p(z) = \frac{1}{N} \sum_{i=1}^N \delta\left(z-f(x_i)\right)
\label{eq:p(z)}
\end{equation}
and the conditional distribution $p(z|x_i)$ as an isotropic normal distribution around the observation in the embedded space. 
\begin{eqnarray}
p(z|x_i) &=& \mathcal{N} ( z | f(x_i), \sigma^2 I) 
\nonumber \\
&=& \frac{1}{\left(2\pi\sigma^2\right)^{d/2}} \exp\frac{\|z-f(x_i)\|^2}{2\sigma^2}
\label{eq:p(z|x_i)}
\end{eqnarray}
where $\sigma$ is the deviation caused by the randomness introduced in DNN training, e.g., batch normalization and dropout layers.

After substituting \eqref{eq:p(z)} and \eqref{eq:p(z|x_i)} into \eqref{eq:I(X;Z)=D_KL}, the derivation is as follows.  
\begin{eqnarray}
&&I(X;Z) = E_{x,z} \left[ \log\frac{p(z|x)}{p(z)} \right] \nonumber \\ 
&=&\int\int \frac{1}{N}\sum_{i=1}^N\delta(z-f(x_i))\log \left[
\frac{1}{\left(2\pi\sigma^2\right)^{d/2}} \exp-\frac{\|z-f(x_i)\|^2}{2\sigma^2}
\right] dz dx\nonumber\\
&&-\int\int\frac{1}{N} \sum_{i=1}^N\delta(z-f(x_i))\log \left[
\frac{1}{N} \sum_{i=1}^N \delta\left(z-z_i\right)
\right] dz dx 
\nonumber\\
&=&
\int \frac{1}{N}\sum_{i=1}^N\log \left[
\frac{1}{\left(2\pi\sigma^2\right)^{d/2}} \exp-\frac{\|z-f(x_i)\|^2}{2\sigma^2}
\right] 
+ \log\frac{1}{N} dx
\nonumber\\
&=& \frac{1}{N^2} \sum_{z\in\mathcal Z}\sum_{i=1}^N\left(-\frac{\|z-f(x_i)\|^2}{2\sigma^2}+\log\sigma^d\right) +\text{const.}
\label{eq:I(X;Z)=E_x,z}
\end{eqnarray}
\eqref{eq:I(X;Z)=E_x,z} interprets as the sum of distances between all pairs of $z$. 

Meanwhile, we model the class conditional probability over $z$ as an isotropic normal distribution around the class center.
\begin{eqnarray}
p(z|y) &=& \mathcal{N}(z; \mu_y, \sigma_yI)
\nonumber\\
&=& \frac{1}{(2\pi\sigma_y^2)^{d/2}}\exp\left(- \frac{\|z-\mu_y\|^2}{2\sigma_y^2} \right)
\end{eqnarray}
where $\sigma_y$ denotes the deviation over the class $y$.

The mutual information $I(Y;Z)$ is then rewritten as follows.
\begin{eqnarray}
I(Y;Z) &= & E_{y,z}\left[\log\frac{p(z|y)}{p(z)}\right]
\nonumber\\
&=& \int\int\frac{1}{N}\sum_{i=1}^N \delta\left(z-z_i\right) \log p(z|y) dy dz
\nonumber\\
&&- \int\int\frac{1}{N}\sum_{i=1}^N  \delta\left(z-z_i\right) \log p(z) dy dz 
\nonumber\\
&=&\int\frac{1}{N}\sum_{i=1}^N \delta\left(z-z_i\right) \log p(z|y) dy
\nonumber\\
&=& \frac{1}{N} \sum_{y=1}^K{n_y}\log \frac{\exp \left(- \frac{\|z-\mu_y\|^2}{2\sigma}\right) -\log\sigma_y^d 
}{\sum \exp \left(- \frac{\|z-\mu_y\|^2}{2\sigma}\right) -\log\sigma_y^d 
}+ \text{const.} 
\label{eq:I(Y;Z)=E_y,z}
\end{eqnarray}
\eqref{eq:I(Y;Z)=E_y,z} is equivalent to the MAP loss function \eqref{eq:L_MCDD} except for the class bias.

To increase $I(Y;Z)$, the intra-class deviation, i.e., the distances between intra-class sample pairs in the embedded space are reduced. Meanwhile, reduction of $I(X;Z)$ is achieved by increasing the distances between inter-class sample pairs. To minimize ${\mathcal L}$, therefore, the intra-class similarity and the inter-class distances  must simultaneously be increased.




\section{Empirical Results}\label{sec:experiments}

\subsection{Setup}
We evaluate the impact of information bottleneck loss in an out-of-distribution detection setting: one designated out-of-distribution class is removed from the training set and a detection measure is computed over the test set \cite{10.1145/3394486.3403189}. 
We conduct the experiment in two steps: (1) train GANs, made up of a generator $G$ and a discriminator $D$, with unlabeled in-distribution samples, and (2) train the embedding layer of the discriminator using the IB loss. 


The set of unlabeled data is denoted as ${\mathcal X}_\text{S} = \{(x_i\}_{i=1}^M$. 
The labeled dataset combines $K-1$ sets of $N$ samples from each class,  ${\mathcal X}_\text{T} = \{\left(p_j, y_j\right)\}_{j=1}^{N\times (K-1)}$.
The label $y_j$ takes a value from ${\mathcal Y}_\text{in}=\left\{1,\ldots,K-1\right\}$.
We us assume here that class $N$ is the OOD class.

The discriminator $D$, which makes prediction between a real and a generated image, is divided into the embedding layers and the fully-connected layers. The former is denoted as an embedding function $f:{\mathcal X}$ with parameters ${\mathcal W}$, 
which maps the input image ${\mathcal X}$ to a deep feature space ${\mathcal Z}$. 

The test set, on which the detection performance is measured, is denoted as ${\mathcal X}_\text{test}=\left\{(x_i, y_i)\right\}_{i=1}^{M_\text{test}}$ where $y_i$ takes a value from ${\mathcal Y}=\{1,\ldots,K\}$. 
The anomaly score of each test sample is computed by kernel density estimation (KDE) in the deep feature space over the labeled examples.

We report the experimental result on the MNIST dataset\cite{726791}, and the AUPRC measure was evaluated for cases where each of $K=10$ class is designated as OOD. The number of labeled samples was set to $N=100$.

\subsection{Results}

\begin{figure}
\centerline{
\includegraphics[width=.7\textwidth]{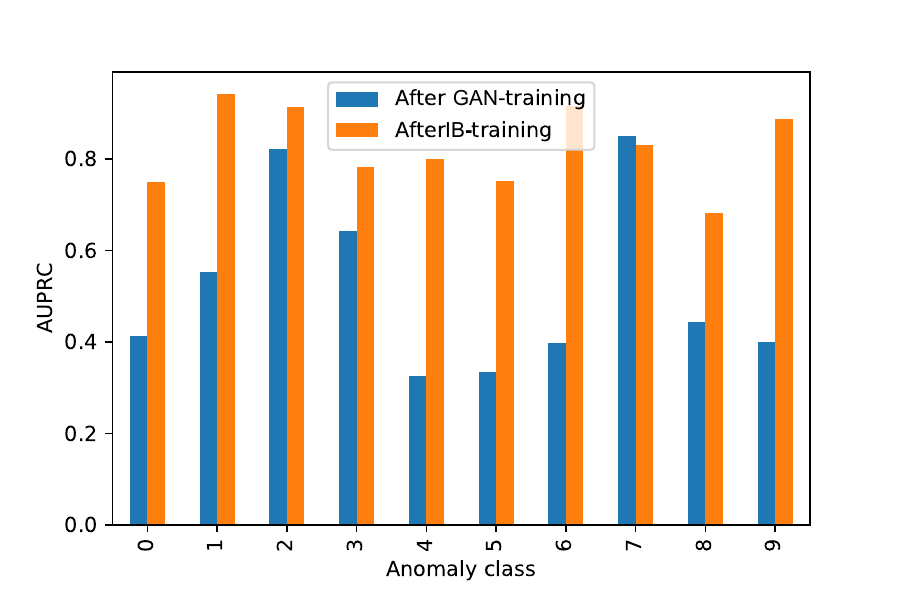}
}
\caption{Comparison in AUPRC (MNIST)}\label{fig:MNIST_barchart}
\end{figure}

\figref{fig:MNIST_barchart} summarizes the AUPRCs in ten settings. The number on the $x$-axis indicate the digit designated as the OOD class. The $y$-axis indicates the mean in ten repetitions. The blue bars indicate the means after step 1 and the orange bar indicate the means after step 2. 
It shows that the training with IB loss yields substantial improvements in AUPRC from those achieved by unsupervised training.

\begin{figure}[htb]
\begin{minipage}{.5\linewidth}
\includegraphics[width=0.99\textwidth]{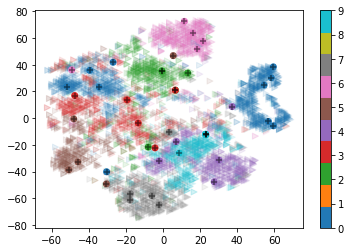}
\caption{After step 1}\label{fig:MNIST_before}
\end{minipage}
\begin{minipage}{.5\linewidth}
\includegraphics[width=0.99\textwidth]{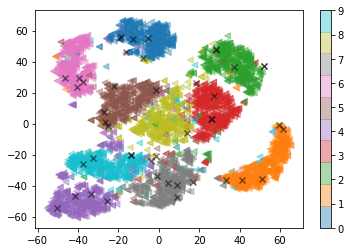}
\caption{After step 2}
\label{fig:MNIST_after}
\end{minipage}
\begin{flushleft}
Low dimensional projection of the embedded features (the OOD class is digit eight). 
\end{flushleft}
\end{figure}

Figures \ref{fig:MNIST_before} and \ref{fig:MNIST_after} illustrate a typical 
low-dimensional projection of the test samples using $t$-SNE \cite{icml2014c2_kusner14} after steps 1 and 2. 
The samples are represented by triangles in colors unique to respective classes while the labeled examples are represented by black $\circ$'s and $\times$'s.
 
From \ref{fig:MNIST_before}, we can see that the normal classes are not distinctly separated after the initial representation learning. \figref{fig:MNIST_after} shows the dispersion of normal classes by IB loss, which contributes to increasing reduce the overlap between in-distribution and OOD classes. 

\section{Conclusion}\label{sec:conclusion}
In this paper, we presented an information-theoretic loss function for training deep neural networks, which takes into account the intra-class similarity and the inter-class discrepancies. The relation between the two terms were derived from the Information Bottleneck principle.
The empirical results indicates that the proposed model can contribute to improved ODD detection performance, from the dispersion of normal classes in the embedded space. 
In future work, we plan to extend this work to a larger collection of tasks such as novelty detection and anomaly detection. 

\newif\ifBibliography\Bibliographyfalse
\ifBibliography
\bibliographystyle{splncs04}
\bibliography{/Users/ando/OneDrive/Bibliography/All.bib}

\begin{thebibliography}{10}
\renewcommand{\url}[1]{}
\providecommand{\urlprefix}{}
\renewcommand{\doi}[1]{}

\bibitem{DBLP:conf/iclr/AlemiFD017}
Alemi, A.A., Fischer, I., Dillon, J.V., Murphy, K.: Deep variational
  information bottleneck. In: 5th International Conference on Learning
  Representations, {ICLR} 2017, Toulon, France, April 24-26, 2017, Conference
  Track Proceedings. OpenReview.net (2017),
  \url{https://openreview.net/forum?id=HyxQzBceg}

\bibitem{1541882}
Chandola, V., Banerjee, A., Kumar, V.: {A}nomaly {D}etection: {A} {S}urvey. ACM
  Comput. Surv.  \textbf{41}(3),  1--58 (2009).
  \doi{http://doi.acm.org/10.1145/1541880.1541882}

\bibitem{goodfellow2016learning}
Goodfellow, I., Bengio, Y., Courville, A.: Deep Learning. MIT Press (2016),
  \url{http://www.deeplearningbook.org}

\bibitem{icml2014c2_kusner14}
Kusner, M., Tyree, S., Weinberger, K.Q., Agrawal, K.: Stochastic neighbor
  compression. In: Jebara, T., Xing, E.P. (eds.) Proceedings of the 31st
  International Conference on Machine Learning (ICML-14). pp. 622--630. JMLR
  Workshop and Conference Proceedings (2014),
  \url{http://jmlr.org/proceedings/papers/v32/kusner14.pdf}

\bibitem{726791}
Lecun, Y., Bottou, L., Bengio, Y., Haffner, P.: Gradient-based learning applied
  to document recognition. Proceedings of the IEEE  \textbf{86}(11),
  2278--2324 (Nov 1998). \doi{10.1109/5.726791}

\bibitem{10.1145/3394486.3403189}
Lee, D., Yu, S., Yu, H.: Multi-class data description for out-of-distribution
  detection. In: Proceedings of the 26th ACM SIGKDD International Conference on
  Knowledge Discovery \& Data Mining. pp. 1362--1370. KDD '20, Association for
  Computing Machinery, New York, NY, USA (2020). \doi{10.1145/3394486.3403189},
  \url{https://doi.org/10.1145/3394486.3403189}

\bibitem{pmlr-v80-ruff18a}
Ruff, L., Vandermeulen, R., Goernitz, N., Deecke, L., Siddiqui, S.A., Binder,
  A., M{\"u}ller, E., Kloft, M.: Deep one-class classification. In: Dy, J.,
  Krause, A. (eds.) Proceedings of the 35th International Conference on Machine
  Learning. Proceedings of Machine Learning Research, vol.~80, pp. 4393--4402.
  PMLR, Stockholmsm{\"a}ssan, Stockholm Sweden (10--15 Jul 2018),
  \url{http://proceedings.mlr.press/v80/ruff18a.html}

\bibitem{DBLP:conf/ipmi/SchleglSWSL17}
Schlegl, T., Seeb{\"{o}}ck, P., Waldstein, S.M., Schmidt{-}Erfurth, U., Langs,
  G.: {Unsupervised Anomaly Detection with Generative Adversarial Networks to
  Guide Marker Discovery}. In: Niethammer, M., Styner, M., Aylward, S.R., Zhu,
  H., Oguz, I., Yap, P., Shen, D. (eds.) Information Processing in Medical
  Imaging - 25th International Conference, {IPMI} 2017, Boone, NC, USA, June
  25-30, 2017, Proceedings. Lecture Notes in Computer Science, vol. 10265, pp.
  146--157. Springer (2017). \doi{10.1007/978-3-319-59050-9\_12},
  \url{https://doi.org/10.1007/978-3-319-59050-9\_12}

\bibitem{Tax:2004:SVD:960091.960109}
Tax, D.M.J., Duin, R.P.W.: Support vector data description. Mach. Learn.
  \textbf{54},  45--66 (January 2004).
  \doi{10.1023/B:MACH.0000008084.60811.49},
  \url{http://portal.acm.org/citation.cfm?id=960091.960109}

\bibitem{7133169}
{Tishby}, N., {Zaslavsky}, N.: Deep learning and the information bottleneck
  principle. In: 2015 IEEE Information Theory Workshop (ITW). pp.~1--5 (2015).
  \doi{10.1109/ITW.2015.7133169}

\bibitem{DBLP:journals/corr/physics-0004057}
Tishby, N., Pereira, F.C., Bialek, W.: The information bottleneck method.
  Computing Research Repository(CoRR)  \textbf{physics/0004057} (2000)

\bibitem{yang2021generalized}
Yang, J., Zhou, K., Li, Y., Liu, Z.: Generalized out-of-distribution detection:
  A survey (2021)

\bibitem{zenati2019efficient}
Zenati, H., Foo, C.S., Lecouat, B., Manek, G., Chandrasekhar, V.R.: Efficient
  gan-based anomaly detection (2019)

\end{thebibliography}
%
\else

\fi

\end{document}